# Extracting health-related causal relations from Twitter messages using Natural Language Processing

Son Doan
Medical Informatics
Kaiser Permanente Southern California
San Diego, CA
Son.Doan@kp.org

Elly W. Yang
Medical Informatics
Kaiser Permanente Southern California
San Diego, CA
Elly.W.Yang@kp.org

Sameer Tilak
Medical Informatics
Kaiser Permanente Southern California
San Diego, CA
Sameer.S.Tilak@kp.org

Manabu Torii
Medical Informatics
Kaiser Permanente Southern California
San Diego, CA
Manabu.Torii@kp.org

*Abstract*—Twitter messages (tweets) concern various types of topics in our daily life, which include health-related topics. Analysis of health-related tweets would help us understand health conditions and concerns encountered in our daily life. In this paper we evaluate an approach to extracting causal relations from tweets using natural language processing (NLP) techniques. Lexico-syntactic patterns based on dependency parser outputs are used for relation extraction. We focused on three health-related topics: "stress", "insomnia", and "headache." A large dataset consisting of 24 million tweets are used. The results show the proposed approach achieved an average accuracy between 74.59% to 92.27% in comparisons to human annotations. Manual analysis on extracted causal tweets reveals interesting findings about expressions on health-related topic posted by Twitter users.

## I. INTRODUCTION

Twitter messages (tweets) are a unique public resource for monitoring health-related information, including, but not limited to, disease outbreaks [1], [2], suicidal ideation[3], [4], obesity [5], and sleep issues [6], [7]. Tweets provide diverse types of information, such as users' behaviors, lifestyles, thoughts, and experiences. This exploratory study focuses on causal relations found in tweets, specifically identifying attributable causes of health problems and concerns. Causal relations in tweets have been studied in the health domain for specific topics, such as adverse reactions caused by drugs [8], [9] or various factors causing stress and relaxation [10]. In this study, we investigate if causes for a given health problem or concern can be extracted from Twitter messages more generally. We focus on three health-related topics: stress, insomnia, and headache.

Text mining from tweets poses various challenges [11]–[13]. One of the challenges in studying causal relations is the small fraction of relevant tweets that need to be accurately spotted in a large data collection. For the simplicity and clarity, in this study, we focus on causal relationships in explicit expressions, such as *"Excessive over thinking leads to insomnia"*. We do not consider implicit or uncertain relationships, such as *"cannot sleep #insomnia #overthinking"*, where "overthinking" is not explicitly stated as the cause of insomnia.

Approaches to this extraction task include identification of frequently co-occurring words or regular expression pattern matching [14]–[16]. These approaches, however, have difficulty in handling complex natural language expressions. For example, co-occurrence-based methods may not be able to distinguish whether the target concept, such as "insomnia" and "stress", is stated as an "effect" or it may be actually stated as a "cause", e.g., *"Insomnia leads to countless thoughts"* and *"This stress is making my chest hurt"*, where "insomnia" and "stress" are the causes. Regular expression-based methods can overcome this issue by explicitly coding phrase occurrence patterns, but it is difficult in practice to cover numerous pattern variations, e.g., *"Overthinking causes headache"*, *"Overthinking causes headache and insomnia"*, *"Overthinking have been frequently causing and worsening headache and insomnia."* Recently, machine learning approaches have been widely used to tackle complex natural language processing (NLP) tasks, including relation extraction tasks [17], [18]. However, training of relation extraction models requires a large amount of hand-annotated data. As an initial exploratory study, a rule-based approach relying on hand-crafted syntactic patterns is a viable and helpful step to understand the problem.

In this study, we created a set of lexico-syntactic patterns to extract "cause" information for a given "effect." We applied our approach to three health-related topics: stress, insomnia, and headache. We used 24 million tweets collected over four months between Sep 30, 2013 and Feb 10, 2014. We manually analyzed the extraction results qualitatively and quantitatively.

## II. BACKGROUND

In the general NLP field, cause-effect relation extraction has been actively studied. There are two main approaches: 1) rule-based methods and 2) machine learning-based methods [16], [17], [19], [20]. Most of these methods use regular expression patterns and apply syntactic patterns to extract and fill in a triple <subject, verb, object>. For example, Cole et al. [20] used an approach to identifying triples of subject, verb, and object, and then applied various rules to determine which of these triples represent causal relations. Blanco et. al [21] used syntactic patterns and then used machine learning to classify those that represent causal relations. Datasets used in these studies are well-written documents, such as news articles and research literature. The reported accuracies vary from F-measures of 0.4 to 0.9, depending on the tasks and dataset. For example, Girju et al. used the lexico-syntactic patterns for extraction, and reported an average of 65% accuracy compared to human annotators [16]. Khoo et al. reported an accuracy of 68% when extracting causal relations from Medline database [22]. A survey on causal relation extraction in the general NLP domain can be found in Asghar et al. [17].

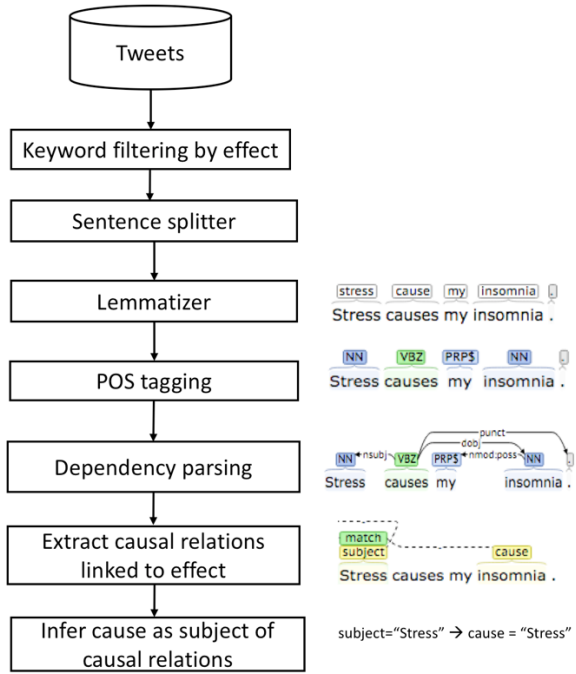

Fig. 1. A general framework to extract causal relations from Twitter messages.

Social media in general and Twitter in particular have been found as a useful and impactful resource in health-related surveillance studies. Twitter data have been used to mine topics related to depression [23], [24], mental health [25], stress and relaxation[10], and tobacco use [26]. Most common techniques for Twitter mining in the health-related domains are keyword look-up and machine learning classification. Support vector machines, logistic regression, and neural networks, among other machine learning algorithms, have been applied in the health domain.

## III. METHODS

### A. Dataset

We used a corpus of 24 million tweets, collected from four cities (New York, Los Angeles, San Francisco and San Diego) over 4-month period (Sep 30, 2013 and Feb 10, 2014). Twitter Streaming API was used to retrieve 1% of all the tweets from these cities during the time period. This corpus was previously used to study stress and relaxation tweets [10]. As the target "effects", we selected three terms: *stress*, *insomnia*, and *headache*.

### B. NLP pipeline

The NLP pipeline for extracting causal relation is shown in Figure 1. First, the corpus is filtered using the target keywords. Next, a series of basic NLP components are applied: sentence splitter, lemmatizer, Part-of-Speech (POS) tagger, and a dependency parser. Finally, causal relations are identified based on syntactic relations generated by the dependency parser. We used CoreNLP package [27] (release version 3.8), a widely used Java library providing various NLP functionalities. The default settings and pre-trained models in the package were used for sentence splitter, lemmatizer, and POS tagger. For the parser, we selected Probabilistic Context-Free Grammar (PCFG) parser and the pre-trained English model in the package, which generates a constituent tree for an input sentence and then converts it into a dependency graph. A dependency graph consists of vertices representing tokens (words and punctuations) and edges representing dependency relations among tokens [28]. Dependency relations are convenient for the purpose of extracting term relations in a sentence. Among several options provided for dependency graph generation in CoreNLP package, we used "Universal Dependencies" (instead of "Original/Stanford Dependencies") and the method generateEnhancedDependencies to derive dependency graphs from parsed trees.

TABLE I. RULE SET TO EXTRACT CAUSAL RELATIONS FROM TWEETS.

| # | Causal relation types | Dependency rules | Examples |
|---|---|---|---|
| 1 | A (noun) caused B | {}=subj < subj ({ + Clausal verb + }=target >dobj {}=cause) | Stress causes insomnia |
| 2 | A (verb-ing) caused B | {}=subj < csubj ({ + Clausal verb + }=target >dobj {}=cause) | Over thinking can increase anxiety and cause insomnia. |
| 3 | B was caused by A | {}=ncsubjpass<nsubjpass({ + Clausal verb + }=target >/nmod:agent/ {}=cause) | My insomnia was caused by stress. |
| 4 | A is a reason of B | Clausal noun + < nsubj ({}=target > /nmod:of/ {}=cause) | Stress is a reason of my insomnia |
| 5 | B was caused by A (verb-ing) | {}=nsubj< nsubjpass ({}=target > /advcl:by/ + Clausal noun ) | Insomnia was caused by overthinking |
| 6 | A results "in/to/from" B | Clausal verb + <[nc]subj ({}=target> /nmod:(to|in|from)/ {}=cause) | Stress results to insomnia. |

## C. Cause-Effect Relation Extraction

There are many different ways to state cause and effect relations, including verb phrases and noun phrases. In order to extract cause-effect relation, we created a rule set template including clausal verbs, clausal verb phrases, and clausal noun phrases. For example, a tweet containing "A caused B" has "caused" as a clausal verb, or "A result to B" has "result to" as a clausal verb phrase. Specifically, the cause-effect relations are determined by the clausal verbs, clausal verb phrases, and clausal noun phrases as below:

*Clausal verb*: "cause, stimulate, make, derive, trigger, result, lead". It is used to identify the relation "A cause B", e.g., "Stress caused insomnia".

*Clausal verb phrase*: "cause, result, reason" + preposition (in, to, from). It is used to is to detect relation "A result to B", e.g., such as "Stress results to insomnia".

*Passive clausal verb*: Past participle of clausal verb + "by". For example, "caused by", "trigged by". It is used to detect relation "A is cause by B", e.g. "Stress was caused by insomnia".

*Clausal noun phrase*: "cause, result, reason" + "of". It is used to detect relation "A is a result of B", e.g., "Insomnia is a result of stress".

We created a set of six general rules to identify cause-effect relationship from verb and noun phrase as above. Those rules are based on syntactic relations derived from a dependency graph generated by a dependency parser. We used CoreNLP Semgrex [29]. Semgrex facilitates subgraph pattern matching over a dependency graph. The details of rules and their examples are listed in Table I. For example, Rule 1 in Table I "{}=subj <subj ({word: /cause/}=target >dobj {}=cause)" indicates that the rule will match a sentence, such as "Stress caused my insomnia", where "Stress" is matched by the pattern "{}=subj" and "insomnia" is matched with the pattern "{}=cause." (Figure 1).

The final step is to extract causes from extracted cause-effect relations. To do so, we extracted the triple <cause, relation, effect>, where effect is one of the three health-related topics of our focus: insomnia, stress and headache.

## IV. RESULTS

We observed that the number of tweets containing specific health-related cause-effect relationships is small in comparison to the overall corpus. Specifically, the number of matching rules is 501 from 29705 (1.6%) tweets for stress, 72/3827 (1.8%) for insomnia, and 94/11252 (0.8%) for headache, respectively. The final causal relationships extracted are 41, 98 and 42 for insomnia, stress and headache. The details of matching rules and number of extracted causal relationship are shown in Table II.

TABLE II. RESULTS WHEN APPLYING RULE SET IN TABLE I TO A CORPUS OF 24 MILLIONS TWEETS. THE LAST ROWS INDICATES THE NUMBERS OF TWEETS EXTRACTED WITH GIVEN EFFECTS (INSOMNIA, STRESS AND HEADACHE).

| MATCHED RULE # | Insomnia (of 3827) | Stress (of 29705) | Headache (of 11252) |
|---|---|---|---|
| 1 | 58 | 381 | 78 |
| 2 | 4 | 12 | 3 |
| 3 | 0 | 4 | 1 |
| 4 | 1 | 21 | 2 |
| 5 | 0 | 32 | 0 |
| 6 | 9 | 51 | 10 |
| Total | 72 | 501 | 94 |
| # causal relationship | 41 | 98 | 42 |

Table II also indicates that the majority pattern of cause-effect relation in tweets is "A caused B" (Rule 1), followed by 'A results "in/to/from" B' (Rule 6). The remaining rules have much smaller portions. This may suggest that Twitter users generally prefer direct and concise expressions. Notably, similar or the same phrases are repeated in collected tweets. For example, similar phrases *"missing someone causes insomnia"*, *"missing someone often causes insomnia"*, and *"missing someone causes insomnia like symptoms"* are found.

### A. Qualitative analysis

To evaluate the accuracy of causal relation extraction, we compared the system outputs to human annotations. Three human annotators [SD, EY, MT] discussed and annotated the system outputs. We annotated at two levels: strict annotation and relaxed annotation. With strict annotation, extracted relations are considered correct only when the cause of the target effect is clearly and explicitly stated. In relaxed annotation, negated or hypothetical statements are additionally considered as correct extraction. For example, *"Cell phone radiation can cause insomnia"*, where the statement is considered hypothetical, is annotated as a false positive case in strict annotation, but a true positive case in relaxation. The disagreement in annotation were resolved by discussions among the annotators.

TABLE III. ACCURACY OF EXTRACTED CAUSAL RELATIONS WHEN COMPARING TO HUMAN ANNOTATORS.

|  | Strict evaluation | Relax evaluation (exclude hypothetical and negation) |
|---|---|---|
| **Insomnia** | 73.81% | 88.10% |
| **Stress** | 82.65% | 96.94% |
| **Headache** | 56.10% | 85.37% |
| Micro-average | 74.59% | 92.27% |

We use accuracy to measure the correctness of the system. The accuracy is calculated by the number of true positives annotated by human annotators divided by the number of tweets system found. The micro-average is calculated by a sum of all true positives across all three categories divided by the total tweets reviewed.

Table III shows the accuracy when comparing system outputs to human annotations. It shows that the micro-average for strict and relaxation is 74.59% and 92.27%, respectively. It also indicates that finding causal relationships for "headache" is more difficult than "insomnia" and "stress". The large variations of strict and relaxation evaluation (74.59% vs. 92.27%) also indicates that hypothetical and negation play important roles in determining causal relationships in Twitter messages.

### B. Quantitative analysis

We further manually analyzed the causes of insomnia, stress and headache extracted by the system. Below are several findings.

*Insomnia*

We found that most frequent causes relating to insomnia is about "missing someone". Other causes include overthinking, social media (Facebook, Twitter), hunger. Below are some examples of tweets and matched rules extracted from this topic:

```
Missing someone causes insomnia.
RULE 1: someone/NN...causes/VBZ ...insomnia/NN

Night before first day of school always results in
insomnia.
RULE 6: Night/NN...results/VBZ ...insomnia/NN
```

*Stress*

The main topics Twitter users talking about stress included school, money, emails, computer games, and physical pains. Below are some of examples and matched rules for this topic.

```
Money only causes stress and conflict
RULE 1: Money/NN...causes/VBZ ...stress/NN

School is the main cause of my stress
RULE 4: School/NNP...cause/NN ...stress/NN
```

*Headache*

We observed the causes of headache Twitter users talking include people, stress, crying, listening. Below are some examples:

```
My neck just made my headache 100x worse
RULE 1: neck/NN...made/VBD ...headache/NN

Nervous Stressed Leads to swollen eye & headaches
RULE 6: Nervous/JJ...Leads/VBZ ...headaches/NNS

You're the cause of my headaches.
RULE 4: You/PRP...cause/NN ...headaches/NNS

too many tears leads to headaches and heavy hearts
RULE 6: tears/NNS...leads/VBZ ...headaches/NNS
```

## V. DISCUSSION

Identifying target tweets precisely and efficiently is a primary key in mining Twitter messages, which contains very large data and time-sensitive information. The goal in our experiment was to identify correctly tweets referring to causal information in a large data set. A dependency parser and associated NLP techniques were used to help improve precise information extraction.

We manually reviewed tweets identified by the proposed approach. We observed that the number of extracted causal relationship tweets is small. However, evaluation showed that it achieved high accuracy. This indicates that using lexicon-syntactic relations derived from dependency parser yields high precision which is an important factor when mining from large data set.

*Limitations.* The study has several limitations. First, in this study, we consider a simple case of causal relation which indicates within one sentence only. In reality, there may have cause-effect relation between different sentences or tweets. Second, in Twitter message there are several ways to imply the cause-effect relation including hashtags or implicit expressions. Third, the data we used in this study is small with 1% sampled from the real data.

## VI. CONCLUSIONS

In this paper, we presented an NLP approach to extracting cause-effect relationships from Twitter messages. The results on four months Twitter data revealed some interesting findings about different health-related topics. In the future we will focus more on semantic analysis such as hashtags as well as multi-sentence causal relation extractions from tweets.